\documentclass{article}

    \usepackage[preprint]{neurips_2024}

\usepackage{packages/packages}
\usepackage{packages/macros}

\title{\fontsize{16}{\baselineskip}\selectfont How Fast Can I Run My VLA? \\Demystifying VLA Inference Performance with \textit{VLA-Perf}}

\author{%
  Wenqi Jiang \\
  NVIDIA Research \\
  \And
  Jason Clemons \\
  NVIDIA Research \\
  \And
  \And
  Karu Sankaralingam \\
  NVIDIA Research \\
  \And
  \And
  Christos Kozyrakis \\
  NVIDIA Research \\
}

\makeatletter
\newcommand{\githublink}[1]{%
  \stepcounter{footnote} 
  \let\old@footnotemark\footnotemark%
  \renewcommand{\footnotemark}{}%
  \footnotetext{#1}%
  \let\footnotemark\old@footnotemark%
}
\makeatother

\begin{document}
\maketitle
\githublink{Code available at: \url{https://github.com/NVlabs/vla-perf}.}


\begin{abstract}
Vision-Language-Action (VLA) models have recently demonstrated impressive capabilities across various embodied AI tasks.
While deploying VLA models on real-world robots imposes strict real-time inference constraints, the inference performance landscape of VLA remains poorly understood due to the large combinatorial space of model architectures and inference systems.
In this paper, we ask a fundamental research question: \textit{How should we design future VLA models and systems to support real-time inference?}
To address this question, we first introduce \ours, an analytical performance model that can analyze inference performance for arbitrary combinations of VLA models and inference systems.
Using \ours, we conduct the first systematic study of the VLA inference performance landscape.
From a model-design perspective, we examine how inference performance is affected by model scaling, model architectural choices, long-context video inputs, asynchronous inference, and dual-system model pipelines.
From the deployment perspective, we analyze where VLA inference should be executed --- on-device, on edge servers, or in the cloud --- and how hardware capability and network performance jointly determine end-to-end latency.
By distilling 15 key takeaways from our comprehensive evaluation, we hope this work can provide practical guidance for the design of future VLA models and inference systems.

\end{abstract}

\section{Introduction}

Embodied AI is widely regarded as a promising next phase of AI, with the potential to enable physical agents that can perceive, reason, and act in the real world. 
Notably, Vision–Language–Action (VLA) models have recently demonstrated strong capabilities in general-purpose manipulation tasks by integrating visual perception and language understanding into the action generation process~\cite{zitkovich2023rt, black2024pi0, intelligence2504pi0, amin2025pi, team2025gemini, bjorck2025gr00t}.

To react to real-time changes in the physical world, VLA inference must operate with low latency, motivating recent work to treat inference performance\footnote{In this paper, \textit{performance} always refers to inference latency and throughput, rather than task success rate.} as a first-class concern in VLA model design.
Such efforts include adopting smaller models~\cite{wen2025tinyvla, shukor2025smolvla, lin2025evo} and quantization~\cite{kim2024openvla, wang2025bitvla}, skipping selected layers~\cite{yue2024deer, yangdysl}, using fewer denoising steps in diffusion-based models~\cite{bjorck2025gr00t}, enabling asynchrony between model inference and robot execution~\cite{black2025real, black2025training, sendai2025leave, tang2025vlash}, and adopting dual-system designs comprising two models of different scales, where only the smaller model operates at high frequency~\cite{figure2025helix, zhang2024hirt, song2025hume}.

While these efforts on efficient VLA model design are an important step forward, we still lack a comprehensive understanding of the VLA inference performance landscape, which is determined by the vast combinatorial space of possible (1) \textit{models} and (2) \textit{inference systems}.
Here, an \textit{inference system} is a combination of (a) the inference accelerator, ranging from edge GPUs to datacenter-class GPUs; (b) the location where inference is executed --- on device, on server, or hybrid; and (c) for server-side inference, the wired or wireless network connecting the robot and the server.
As we will show in our evaluation, executing the same VLA model across different inference systems can lead to performance differences of multiple orders of magnitude.

In this paper, \textbf{we present the first systematic study of VLA inference performance.}
This study aims to answer a simple yet fundamental question: \textit{how should we design VLA models and systems to achieve real-time inference performance?}
Given that standard RGB camera frame rates typically range from 24 to 60~Hz, we define a 10~Hz inference frequency as \textit{acceptable} (not too far from video ingestion rates) and 100~Hz as \textit{high-performance} (exceeding common ingestion rates).
Based on this assumption, we further break down our research question to a series of concrete questions:

(1) From the perspective of VLA models, we ask: \textbf{how should future VLA models be designed under real-time performance constraints?} 
In particular, how much can we scale up model sizes while achieving real-time inference~(\S\ref{sec:eval:scaling})?
Are long-context VLAs that possess thousands of visual frames practically feasible~(\S\ref{sec:eval:longcontext})? 
How does the choice between autoregressive and diffusion-based action experts affect inference performance~(\S\ref{sec:eval:diffusion-vs-ar})? 
How do denoising steps and action chunk size influence performance~(\S\ref{sec:eval:diffusion-steps})?
How much performance gain can be achieved through asynchronous or dual-system inference~(\S\ref{sec:eval:async} and \S\ref{sec:eval:dual_system})?

(2) From a systems perspective, we ask: \textbf{how should we deploy efficient inference systems for various VLA workloads?}
Given a model with verified accuracy, we decompose the deployment problem into the following considerations:
Where should inference be executed --- on device, on server, or via device–server collaboration~(\S\ref{sec:eval:device_vs_server} and \S\ref{sec:eval:device_server_collab})? 
How to choose inference hardware given the various types of available GPUs~(\S\ref{sec:eval:device_vs_server})? 
How critical is network performance in server-side inference systems~(\S\ref{sec:eval:device_vs_server})? 
What combinations of models and systems are required to support VLA inference at rates from 10~Hz up to and beyond 100~Hz~(\S\ref{sec:eval:frequency})?

\begin{figure}[t]
  \centering
  \includegraphics[width=\linewidth]{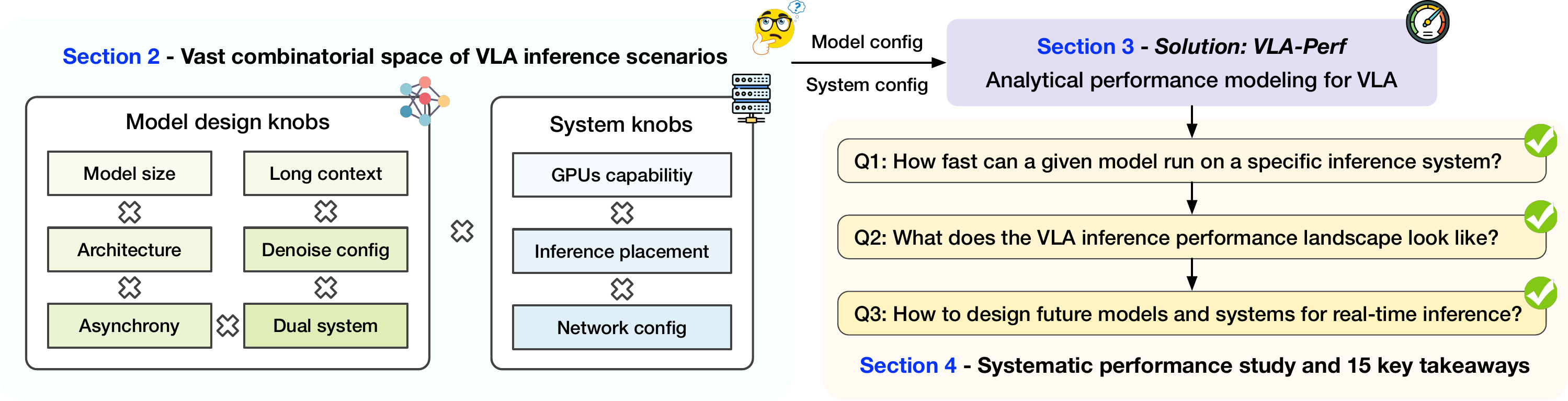}
  \caption{\ours enables a comprehensive performance analysis of the VLA inference landscape. Our systematic study explores the interplay between model architectures and deployment configurations, yielding 15 actionable insights for designing future VLA models and serving systems.}
  \label{fig:overview}
\end{figure}

\textbf{\ours.}
To enable such systematic analysis across the nearly unbounded combinatorial space of VLA models and inference systems, \textit{we develop \ours, an analytical, roofline-based performance model that predicts the optimal inference latency and throughput for arbitrary model–system combinations} (\Cref{fig:overview}).
\ours supports a wide range of VLA configurations, including varying model sizes and architectures, stateless and long-context inference, different action chunk sizes, asynchronous inference, dual-system model pipelines.
In addition, \ours supports diverse deployment scenarios, spanning inference hardware, inference locations, and network configurations.
We open-source \ours to enable further performance analysis beyond those presented in this paper: {\red{\small\url{https://github.com/NVlabs/vla-perf}}}.

Using \ours, we conduct an extensive evaluation of VLA inference performance across a broad space of model variants and system designs.
From the results, we summarize \textbf{15 key performance takeaways that provide practical guidance for the design of future VLA models and inference systems.}

\section{Background and Motivation}
\label{sec:background}

\subsection{Vision-Language-Action Models}
\label{sec:background:vla}

VLA models enable embodied agents to perceive the environment through vision, reason over language instructions, and generate physical actions.
Recent VLA models have demonstrated strong performance on general-purpose manipulation tasks using robotic arms~\cite{brohan2022rt, black2024pi0, kim2024openvla, zhao2023learning} and humanoid robots~\cite{bjorck2025gr00t, figure2025helix}.

\textbf{Model architecture.}
Existing VLA models adopt either \textit{autoregressive-based} or \textit{diffusion-based (including flow matching)} action generation. 
Autoregressive models use a single transformer to integrate visual observations, interpret language instructions, and generate actions, producing one action dimension (or token) at a time in an iterative manner.
Representative examples of this paradigm include the RT series~\cite{brohan2022rt, zitkovich2023rt, gu2023rt}, OpenVLA~\cite{kim2024openvla}, and Octo~\cite{team2024octo}. 
More recently, an alternative VLA paradigm combines a VLM backbone with a separate, typically smaller, diffusion-based action expert. 
Here, the VLM backbone ingests vision and language inputs, while a diffusion-based action expert attends to the VLM’s KV cache and generates actions through an iterative refinement process, with the number of denoising steps~\footnote{For brevity, we use the term \emph{denoising steps} to describe the iterative refinement steps in both classic diffusion models and flow-matching–based variants.} as a configurable parameter.
Representative diffusion-style VLA models include the $\pi_0$ series~\cite{amin2025pi, intelligence2504pi0, black2024pi0}, GR00T~\cite{bjorck2025gr00t}, SmolVLA~\cite{shukor2025smolvla}, and TinyVLA~\cite{wen2025tinyvla}. 

\textbf{Action prediction.}
Each robot action typically consists of multiple dimensions, such as joint positions, velocities, or torques for robotic arms, or whole-body joint configurations for humanoid robots.
To enable smooth and stable execution, many VLA models employ \emph{action chunking}, where the model predicts a sequence of future actions in a single inference~\cite{zhao2023learning, kim2025fine, jing2025mixture}, with the sequence length referred to as the \textit{action chunk size}.
Under action chunking, an \textit{execution horizon} can be specified, defined as the number of actions actually executed before the next inference is performed, which is no larger than the chunk size~\cite{black2025real, black2025training}.
A larger execution horizon can improve action smoothness and reduce inference frequency, but it also reduces the model’s ability to react promptly to changes in the external environment.

\begin{figure}[t]
  \centering
  \includegraphics[width=.9\linewidth]{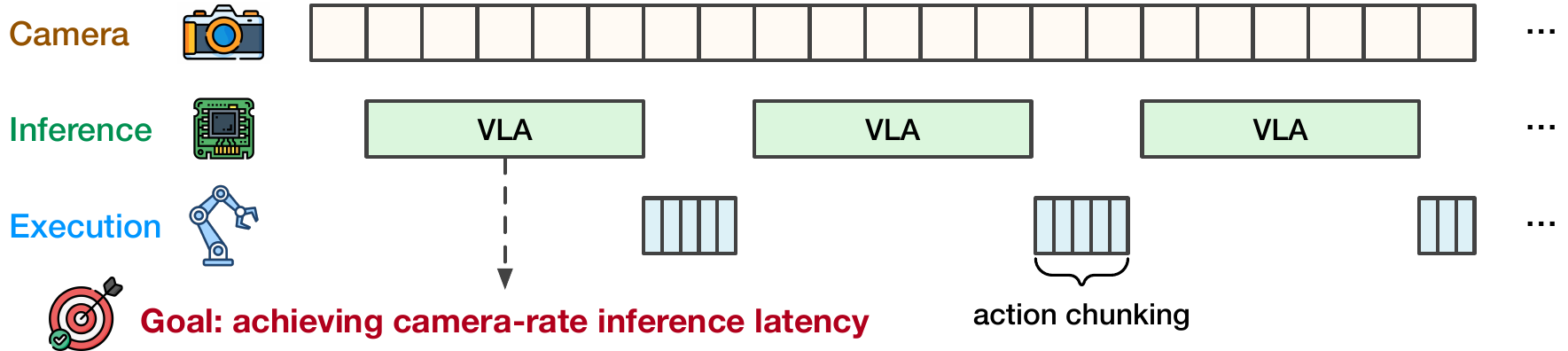}
  \caption{An example timeline of synchronous VLA inference. An efficient inference system should aim to match camera ingest rates to provide the robot with real-time action guidance.}
  \label{fig:background:vla-timeline}
\end{figure}

\subsection{Efficient VLA Inference}
\label{sec:background:efficient_vla}

An VLA system should aim to achieve real-time inference (10 to 100 ms of latency) to match the rate of visual signal ingestion, as visualized in \Cref{fig:background:vla-timeline}.
To meet this demand, a growing number of work has proposed techniques to improve VLA inference efficiency at both the model and system levels~\cite{yu2025survey}.

\textbf{Reduce computation.}
The amount of computation can be reduced by adopting smaller models~\cite{wen2025tinyvla, shukor2025smolvla, lin2025evo, sun2026dadu} and quantization~\cite{kim2024openvla, wang2025bitvla}, skipping selected VLM layers~\cite{yue2024deer, yangdysl}, or reducing the number of denoising steps in diffusion-based action experts~\cite{bjorck2025gr00t}. 
For autoregressive VLAs, parallel decoding can further accelerate inference by reusing KV cache for multi-token predictions~\cite{kim2025fine}. 
Finally, action chunking allows the model to predict a sequence of actions to execute in a single inference call, thereby reducing inference frequency~\cite{zhao2023learning, black2025real}.

\textbf{Asynchronous inference and dual-system VLAs.}
Inference performance can also be improved through various forms of asynchrony.
For example, we can allow inference to begin while the robot is still executing previous actions~\cite{black2025real, black2025training, sendai2025leave, tang2025vlash}.
This inference-execution overlap improves GPU utilization and consequently inference throughput.
Alternatively, a dual-system VLA pipeline runs a lightweight action expert at a higher frequency (System~1) while invoking a more expensive VLM backbone at a lower frequency (System~2)~\cite{figure2025helix, zhang2024hirt, song2025hume}, with the two systems asynchronously exchanging latent states.

\textbf{Better inference systems.}
While higher inference frequencies can be attained through more powerful hardware, software-level optimizations are also critical for VLA inference efficiency.
Careful CUDA-level optimizations, including CUDA graph and operator fusion, can reduce inference latency by up to $5\times$ compared to a naive PyTorch implementation~\cite{ma2025running}.
For server-side inference with action chunking, network latency and robot execution latency can be overlapped to reduce end-to-end execution time~\cite{huang2025dadu}.

\subsection{Research Gap: Comprehensive Analysis of the VLA Inference Performance Landscape}
\label{sec:background:gap}

Despite the advances in efficient VLA designs introduced above, we still lack a comprehensive understanding of the VLA inference performance landscape, largely due to \textit{(1) the wide diversity of inference system configurations across prior studies and (2) the limited exploration of model architectures driven by inference performance concerns.}
From a systems perspective, VLA inference can be performed on edge GPUs integrated within the robot (on-device)~\cite{figure2025helix}, on GPU servers near the robot (edge-server)~\cite{black2024pi0, amin2025pi}, or offloaded to powerful accelerators in the cloud (cloud-server)~\cite{brohan2022rt, zitkovich2023rt} --- the performance across these configurations can vary significantly, as we will show in the evaluation.
From a model-design perspective, while existing VLA models are often designed with inference efficiency in mind~\cite{shukor2025smolvla, wen2025tinyvla}, they often target specific application-system pairings.  
Given the rapid evolution of both workloads and inference hardware, this approach can be myopic and can limit the exploration of alternative designs involving larger models or longer context.

\section{Analyzing VLA Inference Performance with \ours}
\label{sec:methodology}

In this paper, \textbf{we aim to provide a comprehensive analysis of VLA inference performance across both existing and future, potentially hypothetical, combinations of VLA models and inference systems.}
Our evaluation focuses exclusively on performance characteristics --- including latency and throughput --- under the assumption that the underlying models meet the necessary accuracy thresholds for deployment.

However, conducting such a comprehensive analysis is much more challenging than profiling inference performance on a small set of existing models and system implementations.
This is because it requires (1) setting up systems with various accelerator capabilities, inference locations, and network configurations, as discussed in \S\ref{sec:background:gap}, and (2) evaluating not only existing models but also plausible future model variants --- the resulting combinatorial explosion renders exhaustive empirical evaluation both cost- and time-prohibitive.

To address this challenge, \textbf{we adopt a modeling-based approach to performance analysis}, which has shown strong effectiveness in prior work on LLM inference and training systems~\cite{davies2025liminal, bambhaniya2024demystifying, agrawal2024vidur, cho2024llmservingsim, yuan2024llm, jiang2025rago}.
Analytical performance models focus on capturing the dominant performance characteristics of both the model (e.g., FLOPs and memory accesses) and the hardware (e.g., peak FLOP/s, memory bandwidth, and network bandwidth), and estimate achievable performance based on these attributes.
This approach enables fast, low-cost performance analysis across arbitrary combinations of models and hardware, without requiring the deployment of real systems.
On the downside, analytical models are not perfectly accurate, as they typically assume optimistic software implementations and therefore estimate the upper bound of achievable performance.
For example, a recent study on optimizing VLA inference performance reports that 68$\sim$75\% of roofline-model-predicted performance can be achieved on real systems~\cite{ma2025running}.
While such predictions are not exact, we are still at an early stage in understanding VLA inference performance, and thus even coarse-grained estimates can provide valuable guidance for future model and system designs.

\begin{figure}[t]
  \centering
  \includegraphics[width=\linewidth]{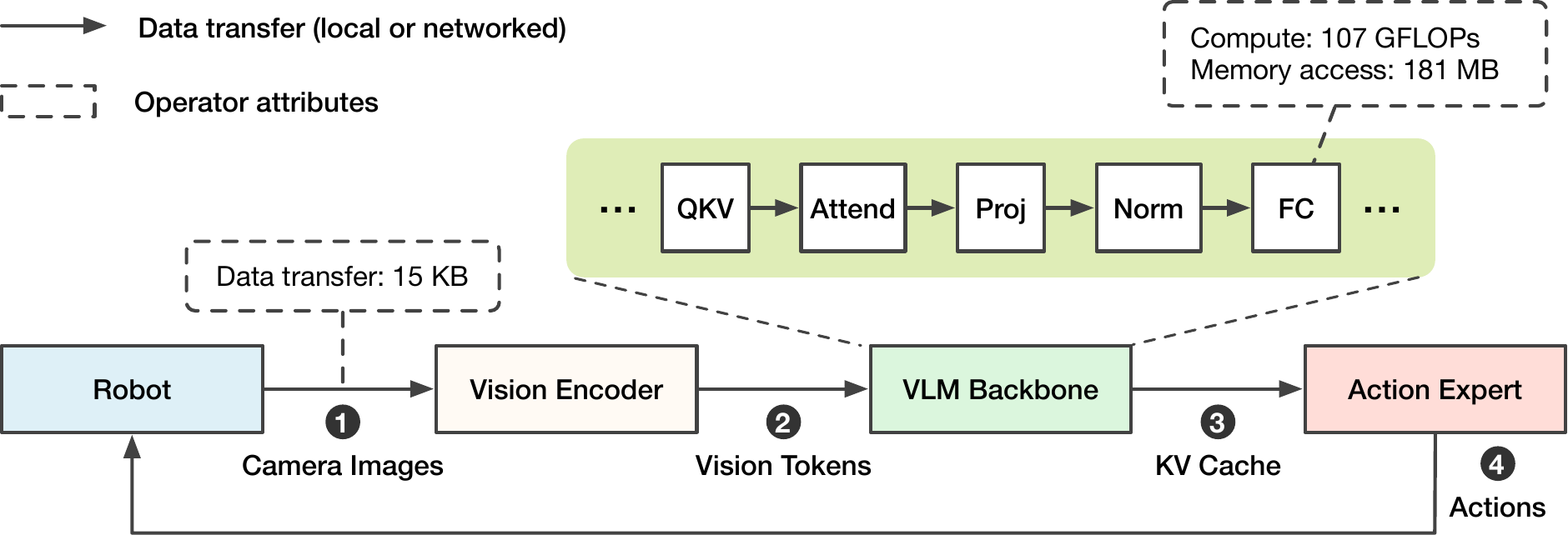}
  \caption{\ours abstracts VLA inference as model components interleaved with data transfers.}
  \label{fig:vla-perf-abstraction}
\end{figure}

\textbf{\ours overview.}
We build \ours, a roofline-based analytical performance model for VLA inference.
Figure~\ref{fig:vla-perf-abstraction} illustrates an example VLA inference workflow, which consists of a robot and multiple model components.
Depending on the placement of each model component (either on the robot or on a server), these components exchange data either locally or over a network.
Such data transfers may include raw images, vision tokens, KV caches, or action predictions.
Each model component is abstracted as a sequence of operators, such as fully connected layers, linear projections, and attention blocks.
\ours assumes that inference for each individual model component (e.g., the VLM backbone) is executed on a single accelerator, because modern GPUs, including recent edge accelerators, already provide sufficient memory capacity to host complete VLA models, for example up to 128~GB on NVIDIA Jetson Thor.
On the contrary, different model components can be executed either on the same accelerator or on different accelerators.

\begin{figure*}[t] %
    \centering
    \begin{minipage}[t]{0.48\textwidth}
        \centering
        \textbf{VLA Model Configuration}
\begin{lstlisting}
# pi0 VLM Backbone (Gemma 2B)
pi0_vlm = ModelConfig(
    seq_len=800, # language + 3 images 
    hidden_size=2048,
    intermediate_size=16384,
    num_ffi=2,
    num_decoder_layers=18,
    num_attention_heads=8,
    head_dim=256,
)

# pi0 Action Expert (diffusion-based)
pi0_action_expert = ModelConfig(...)

# pi0 Vision Encoder (SigLIP)
pi0_vision_encoder = ModelConfig(...)

\end{lstlisting}
    \end{minipage}
    \hfill
    \begin{minipage}[t]{0.48\textwidth}
        \centering
        \textbf{Inference System Configuration}
\begin{lstlisting}
# GPU Capability, e.g., NVIDIA B100
GPU_CONFIG = AcceleratorConfig(
    name='B100',
    BF16_TFLOPS=1750,
    Memory_GB'=192,
    HBM_BW_GBs=8000,
    ...
)

# Network Environment, e.g., Ethernet 1G
NET_CONFIG = NetworkConfig(
    name='Ethernet 1G',
    bandwidth_mbps=1000,
    base_latency_ms=0.1,
    efficiency=1.0
)

\end{lstlisting}
    \end{minipage}
    \caption{Example inputs to \ours, including model parameters (left) and system specifications (right).}
    \label{fig:methodology:input_params}
\end{figure*}

\textbf{Input parameters.}
\ours enables the analysis of arbitrary model-system combinations by parameterizing both model and system parameters, with an example provided in~\Cref{{fig:methodology:input_params}}.
On the model side, these parameters include the choice of vision encoder, VLM backbone, and action expert; the input and output sequence lengths of each model; the number of denoising steps for diffusion-based action experts; action chunk size; and the dimensionality of each action.
On the system side, \ours supports various inference accelerators of configurable peak FLOP/s and memory bandwidth, as well as network systems characterized by upload/download bandwidth and latency.

\textbf{Latency calculation.}
Given the inputs above, the end-to-end inference latency of a VLA system is modeled as the sum of model inference latency and data movement latency across all components:
\begin{equation}
T_{\text{total}} = \sum_{m \in \mathcal{M}} T_m + \sum_{d \in \mathcal{D}} T_d ,
\end{equation}
where $\mathcal{M}$ denotes the set of model inference components and $\mathcal{D}$ denotes the set of data movement stages.

For a single model component $m$, the inference latency $T_m$ is modeled as the sum of latency of each of its constituent operators:
\begin{equation}
T_m = \sum_{o \in \mathcal{O}_m} T_o ,
\end{equation}
where $\mathcal{O}_m$ denotes the sequence of operators in model $m$.
For each operator $o$, \ours models its execution latency using a roofline model that accounts for both compute and memory access latency:
\begin{equation}
T_o = \max \left(
\frac{\text{FLOPs}_o}{\text{FLOP/s}_h}, 
\frac{\text{Bytes}_o}{\text{MemBW}_h}
\right) ,
\end{equation}
where $\text{FLOPs}_o$ and $\text{Bytes}_o$ denote the total floating-point operations and memory bytes accessed by operator $o$, while $\text{FLOP/s}_a$ and $\text{MemBW}_a$ denote the peak compute throughput and memory bandwidth of the inference hardware $h$, respectively.

We assume that local data movement on the same accelerator is sufficiently fast to be treated as negligible, while network-based data movement between devices is modeled as:
\begin{equation}
T_d^{\text{net}} = \text{NetLat} + \frac{\text{Bytes}_d}{\text{NetBW}} ,
\end{equation}
where $\text{Bytes}_d$ denotes the amount of transferred data, and $\text{NetBW}$ and $\text{NetLat}$ denote the single-directional network bandwidth and latency, respectively.

\begin{table}[t]
    \centering
    \footnotesize 
    \setlength\dashlinedash{0.2pt}
    \setlength\dashlinegap{1.5pt}
    \caption{Roofline model validation against real $\pi_0$ Triton inference latencies on an RTX 4090~\cite{ma2025running}. This evaluation uses 10 flow-matching steps, action chunk size of 63, and an empty language prompt.}
    \vspace{.5em}
    \label{tab:validate_perf_4090}
    \begin{tabular}{L{10em} R{5em} R{5em} R{5em}}
\toprule
\textbf{Metric} & \multicolumn{1}{c}{\textbf{1 camera}} & \multicolumn{1}{c}{\textbf{2 cameras}} & \multicolumn{1}{c}{\textbf{3 cameras}} \\
\midrule
Roofline (\ours)    & 14.7 ms & 22.5 ms & 30.4 ms \\
\hdashline
Real Perf. (Triton) & 20.0 ms & 27.3 ms & 36.8 ms \\
\hdashline
Fidelity (Real/Roofline)            & \cellcolor{lightgray} 73.3\% & \cellcolor{lightgray} 82.3\% & \cellcolor{lightgray} 82.6\% \\
\bottomrule
\end{tabular}
\end{table}

\textbf{Modeling fidelity.}
Due to the scarcity of well-optimized frameworks for VLA inference, we mainly validate the fidelity of \ours using the $\pi_0$ implementation by Ma et al.~\cite{ma2025running}, a Triton-based implementation specifically tuned for RTX 4090.
\Cref{tab:validate_perf_4090} compares the performance predicted by \ours to the empirical measurements conducted by Ma et al.~\cite{ma2025running}. 
The results demonstrate that an optimized system can achieve $73.3\sim82.6\%$ of the theoretical roofline reported by \ours, with the gap narrowing as the workload increases (e.g., when processing three camera frames).

The performance differences between a real inference system and the roofline limits reported by \ours are due to both hardware and software factors. 
First, \ours abstracts away hardware-specific details, including microarchitectural design, instruction scheduling, and memory-access behavior. 
Instead, \ours assumes that the maximum theoretical compute capability and memory bandwidth are attainable for every operator executed. 
Second, real-world systems incur software overheads, such as kernel launch latencies, operating system interference, and runtime library overhead, which are not explicitly modeled by \ours.
Nevertheless, we believe that a modeling fidelity exceeding 80\% is sufficient to provide meaningful insights into the VLA performance landscape, and thus leave the tuning of \ours for specific hardware and software platforms, such as in~\cite{davies2025liminal, jiang2025rago}, to future work.

\section{Evaluation and Takeaways}
\label{sec:evaluation}

In this section, we use \ours to conduct a comprehensive analysis of VLA inference performance.
Our evaluation is structured to address two sets of research questions as below.

\textbf{Question 1: How should we design future VLA models to meet real-time latency constraints?}
\vspace{-.5em}
\begin{itemize}[leftmargin=1.2em]
    \item How far can model sizes be scaled while still enabling real-time inference~(\S\ref{sec:eval:scaling})?
    \item Are long-context VLAs that process thousands of visual frames practically feasible~(\S\ref{sec:eval:longcontext})?
    \item How do autoregressive and diffusion-based action experts compare in performance~(\S\ref{sec:eval:diffusion-vs-ar})?
    \item How do denoising steps and action chunk size influence performance~(\S\ref{sec:eval:diffusion-steps})?
    \item Are asynchronous or dual-system inference much faster than synchronous inference~(\S\ref{sec:eval:async} and \S\ref{sec:eval:dual_system})?
\end{itemize}

\textbf{Question 2: How should inference systems be deployed for different VLA workloads?}
\vspace{-.5em}
\begin{itemize}[leftmargin=1.2em]
    \item Should inference be executed  on device, on server, or via device-server collaboration~(\S\ref{sec:eval:device_vs_server} and \S\ref{sec:eval:device_server_collab})?
    \item How capable must inference hardware be to meet real-time performance requirements~(\S\ref{sec:eval:device_vs_server})?
    \item How critical is network performance for server-side inference systems~(\S\ref{sec:eval:device_vs_server})?
    \item What model-system combinations can achieve inference rates from 10~Hz to 100~Hz~(\S\ref{sec:eval:frequency})?
\end{itemize}

Our experiments are organized as follows.
\S\ref{sec:eval:baseline} presents a baseline analysis of the $\pi_0$ model.
\S\ref{sec:eval:scaling}$\sim$\ref{sec:eval:diffusion-vs-ar} explore various model configuration to examine their impact on inference performance.
\S\ref{sec:eval:device_vs_server}$\sim$\ref{sec:eval:frequency} additionally considers inference placement and network latency, closely reflecting real-world deployments.

\subsection{Evaluation Setup}
\label{sec:evaluation:setup}

We describe the main model and system settings here, with additional details provided in Appendix~\ref{sec:appendix:parameters}.

\textbf{Models and robot.}
We evaluate a set of model variants derived from the $\pi_0$ architecture~\cite{black2024pi0}, which we choose due to its strong robotic task performance and widespread adoption in recent VLA systems.
The original $\pi_0$ model consists of a 400M SigLIP vision encoder, a 2B Gemma language model, and a 300M diffusion-based action expert.
Throughout the experiments, we consider a bimanual robotic manipulation setting, which is common for both stationary robots and mobile platforms such as wheeled or humanoid robots.
With the UR5e robot arm, this setup is equipped with three cameras and an action space of 14 degrees of freedom (DoF).
Each camera image has a resolution of 224$\times$224 and is tokenized into 256 visual tokens, yielding 768 visual tokens across three cameras.
Assuming 32 language tokens per task, the total input sequence length per inference is 800 tokens.
Unless otherwise specified, we use an action chunk size of 50 and 10 denoising steps for action generation, same as the original $\pi_0$ configuration.

\begin{table*}[t]
    \centering
    \setlength\dashlinedash{0.2pt}
    \setlength\dashlinegap{1.5pt} 
    \caption{We consider systems with various (1) GPU capabilities (rows) and (2) inference location (columns).}
    \label{tab:inference_placement}
    \scalebox{1.0}{
        \begin{tabular}{L{10em} M{6em} M{6em} M{6em}}
\toprule
\textbf{Capacity and Placement}
& \textbf{On-Device}
& \textbf{Edge Server}
& \textbf{Cloud Server} \\
\midrule
Mobile (Thor)
& \checkmark
& 
& \\
\hdashline
Consumer (RTX 4090)
& 
& \checkmark
& \\
\hdashline
Datacenter (B100)
& 
& \checkmark
& \checkmark \\
\bottomrule
        \end{tabular}
    }
\end{table*}

\textbf{Inference systems.}
We evaluate a range of accelerators spanning high-end edge GPUs (e.g., NVIDIA Jetson Thor), consumer-grade GPUs commonly used in research experiments (e.g., RTX 4090), and high-end datacenter GPUs, including A100, H100, and B100.
The GPUs can be mapped to various inference location as shown in~\Cref{tab:inference_placement}.
For server-side inference, we evaluate both wired and wireless network configurations, including Ethernet, WiFi, and cellular (4G/5G) networks.
All experiments assume BF16 for inference, or FP16 when BF16 is not supported by the hardware.

\textbf{Performance metrics.}
We report VLA system latency, defined as the elapsed time from when the robot perceives visual observations to when the robot receives the corresponding action prediction.
We also report throughput in Hertz (Hz), defined as the number of inferences that can be executed per second at a batch size of one (i.e., a single robot).
For synchronous inference, throughput is the inverse of inference latency, whereas for asynchronous inference, throughput can exceed the inverse latency.
We report inference performance independent of robot execution latency, as the latter is highly robot-dependent. 
Furthermore, the effective action execution frequency may exceed the inference frequency due to action chunking --- with a chunk size of five, the robot can execute actions at up to five actions given a single inference.

\subsection{Baseline $\pi_0$ Inference Latency Across Hardware Backends}
\label{sec:eval:baseline}

Before evaluating model and system variants, we first establish a baseline by measuring the inference performance of the $\pi_0$ model across a range of GPUs, without considering network latency.

\begin{table*}[t]
    \centering
    \setlength\dashlinedash{0.2pt}
    \setlength\dashlinegap{1.5pt}
    \begin{footnotesize}
    \caption{Inference performance of $\pi_0$ on various GPUs without considering network latency.}
    \label{tab:pi0_base_performance}
    \scalebox{1.0}{
        \begin{tabular}{L{5em} R{5em} R{5em} R{5em} R{5em} R{5em}}
\toprule
\textbf{Hardware}
& \multicolumn{1}{c}{\textbf{Vision Lat.}}
& \multicolumn{1}{c}{\textbf{VLM Lat.}}
& \multicolumn{1}{c}{\textbf{Action Lat.}}
& \multicolumn{1}{c}{\textbf{E2E Lat.}}
& \multicolumn{1}{c}{\textbf{E2E Freq.}} \\
\midrule
Jetson Thor
& 6.06 ms
& 20.30 ms
& 26.20 ms
& 52.57 ms
& \cellcolor{lightgray} 19.0 Hz \\
\hdashline
RTX 4090
& 4.02 ms
& 19.79 ms
& 7.25 ms
& 31.06 ms
& \cellcolor{lightgray} 32.2 Hz \\
\hdashline
A100
& 2.13 ms
& 10.47 ms
& 3.60 ms
& 16.20 ms
& \cellcolor{lightgray} 61.7 Hz \\
\hdashline
H100
& 0.71 ms
& 3.30 ms
& 2.14 ms
& 6.15 ms
& \cellcolor{lightgray} 162.5 Hz \\
\hdashline
B100
& 0.40 ms
& 1.87 ms
& 0.91 ms
& 3.18 ms
& \cellcolor{lightgray} 314.4 Hz \\
\bottomrule
\end{tabular}
    }
    \end{footnotesize}
\end{table*}
\begin{table}[t]
    \centering
    \setlength\dashlinedash{0.2pt}
    \setlength\dashlinegap{1.5pt}
    \begin{footnotesize}
    
    \caption{Compute- vs. memory-bound analysis of $\pi_0$ across different hardware. Operator intensity (OI) denotes the ratio between compute operations and memory accesses (FLOPs/Bytes). The balance OI denotes the hardware balance point at which compute throughput and memory bandwidth are equally limiting.}

    \vspace{.5em}
    \label{tab:pi0_base_workload}
    \scalebox{1.0}{
        \begin{tabular}{L{5em} M{5em} M{6em} M{6em} M{6em}}
\toprule
\textbf{Hardware}
& \multicolumn{1}{c}{\textbf{Balance OI}} 
& \multicolumn{1}{c}{\textbf{Vision} (OI=321.4)}
& \multicolumn{1}{c}{\textbf{VLM} (OI=542.8)}
& \multicolumn{1}{c}{\textbf{Action} (OI=54.0)} \\
\midrule
Jetson Thor 
& 1481.5 
& Memory
& Memory
& Memory \\
\hdashline
RTX 4090 
& 163.7 
& Compute
& Compute
& Memory \\
\hdashline
A100 
& 153.0 
& Compute
& Compute
& Memory \\
\hdashline
H100 
& 295.2 
& Compute
& Compute
& Memory \\
\hdashline
B100 
& 218.8 
& Compute
& Compute
& Memory \\
\bottomrule
\end{tabular}
    }
    \end{footnotesize}
\end{table}

\textit{\uline{Takeaway 1: Existing datacenter GPUs can already achieve inference frequencies comparable to camera frame rates for small VLA models such as $\pi_0$, while edge GPUs remain performance-limited.}}

\Cref{tab:pi0_base_performance} shows that A100, H100, and B100 achieve inference frequencies ranging from 61.7~Hz to 314.4~Hz, which are at least on par with the frame rates of common RGB cameras (24$\sim$60~Hz).
In contrast, Jetson Thor achieve substantially lower inference frequency (19.0~Hz), falling below the frame rates of most cameras.

\textit{\uline{Takeaway 2: Action prediction is memory-bound across hardware, while vision and VLM inference are compute-bound on most GPUs except from Jetson Thor.}}

\Cref{tab:pi0_base_workload} summarizes the workload characteristics of each VLA model component.
The vision encoder and the VLM backbone exhibit significantly higher operator intensity (321.4 and 542.8~FLOPs/Byte, respectively) compared to the action expert (54.0~FLOPs/Byte).
This is because the vision encoder and the VLM backbone process many input tokens (e.g., 768 for SigLIP and 800 for Gemma), inherently batching computation across tokens, whereas the diffusion-based action expert operates on far fewer tokens (e.g., the same as the action chunk size of 50).
This behavior closely mirrors LLM inference, where the prefill phase (prompt processing) is compute-intensive, while the decode phase (token generation) is memory-bound~\cite{patel2024splitwise}.
Jetson Thor, in contrast to the other evaluated GPUs, relies on LPDDR memory, which prioritizes low power consumption for embedded devices but provides substantially lower bandwidth (270~GB/s) than GDDR on RTX~4090 (1~TB/s) and HBM on B100 (8~TB/s).
As a result, even the vision encoder and VLM backbone become memory-bound on Jetson Thor.

\subsection{Scaling Model Sizes Under Real-Time Constraints}
\label{sec:eval:scaling}

We next study how inference latency scales with increasing VLA model sizes, which are positively correlated with task accuracy~\cite{generalist2025gen0}.
Specifically, we scale each component of the $\pi_0$ model and construct a family of larger VLA models.
For the vision encoder, we replace the original SigLIP-So400m used in $\pi_0$ with the larger SigLIP-Giant model with 1.1B parameters.
For the VLM, we replace Gemma with the Llama2 family (7B, 13B, and 70B), which provides a wider range of model scales.
For the action expert, we follow the $\pi_0$ design principle and instantiate it as a scaled-down version of the corresponding VLM, with approximately $4\sim8\times$ fewer parameters by reducing the transformer hidden dimension and intermediate dimension by $2\times$ and $4\times$, respectively.
By combining these components, we construct a set of hypothetical larger VLA models, denoted as $\pi_0$-L, $\pi_0$-XL, and $\pi_0$-XXL, whose configurations are summarized in \Cref{tab:pi0_model_scaling}.

\begin{table*}[t]
    \centering
    \setlength\dashlinedash{0.2pt}
    \setlength\dashlinegap{1.5pt}
    \caption{Inference performance of scaled-up VLA models across different hardware platforms.}
    \label{tab:pi0_model_scaling}
    \scalebox{0.76}{
\begin{tabular}{@{} L{6.5em} R{8em} R{8em} R{7em} R{5em} R{4.5em} R{4.5em} @{}}
\toprule
\textbf{Model}
& \multicolumn{1}{c}{\textbf{Vision Encoder}}
& \multicolumn{1}{c}{\textbf{VLM}}
& \multicolumn{1}{c}{\textbf{Action Expert}}
& \multicolumn{1}{c}{\textbf{Jetson Thor}}
& \multicolumn{1}{c}{\textbf{RTX 4090}}
& \multicolumn{1}{c}{\textbf{B100}} \\
\midrule
$\pi_0$ (2.7B)
& SigLIP-So (0.4B)
& Gemma-2B (2.0B)
& Act-M (0.3B)
& \cellcolor{lightgray}19.0 Hz
& \cellcolor{lightgray}32.2 Hz
& \cellcolor{lightgray}314.4 Hz \\
\hdashline
$\pi_0$-L (9.1B)
& SigLIP-Giant (1.1B)
& Llama2-7B (6.5B)
& Act-L (1.5B)
& \cellcolor{lightgray}3.9 Hz
& \cellcolor{lightgray}8.0 Hz
& \cellcolor{lightgray}73.6 Hz \\
\hdashline
$\pi_0$-XL (16.7B)
& SigLIP-Giant (1.1B)
& Llama2-13B (12.7B)
& Act-XL (2.9B)
& \cellcolor{lightgray}2.1 Hz
& N/A
& \cellcolor{lightgray}39.7 Hz \\
\hdashline
$\pi_0$-XXL (81.3B)
& SigLIP-Giant (1.1B)
& Llama2-70B (68.5B)
& Act-XXL (11.7B)
& N/A
& N/A
& \cellcolor{lightgray}9.6 Hz \\
\bottomrule
\end{tabular}
}
\end{table*}

\begin{figure}[t]
  \centering
  \includegraphics[width=\linewidth]{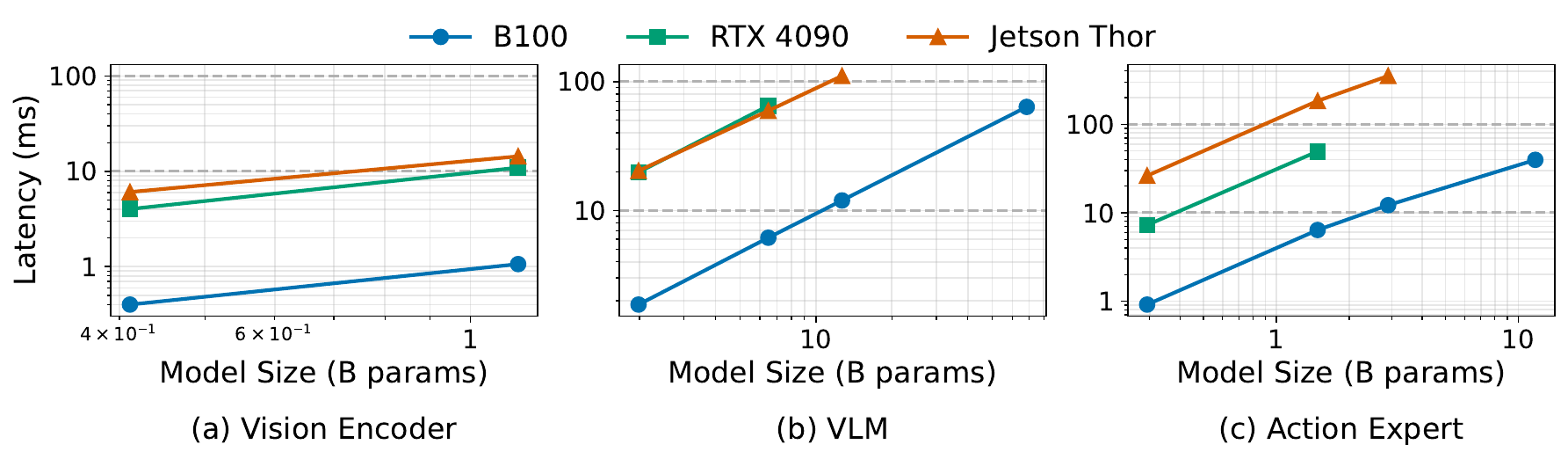}
  \vspace{-1em}
  \caption{Increased model sizes lead to proportional inference latency increases.}
  \label{fig:eval:model_scale}
\end{figure}

\textit{\uline{Takeaway 3: Latency of each VLA component scales approximately linearly with increasing model sizes.}}

\Cref{fig:eval:model_scale} breaks down the inference latency of individual VLA components as model size increases, where both axes are shown on a logarithmic scale.
Across all components, larger models impose proportionally higher computational costs, and thus inference latency grows approximately linearly with model size.

\textit{\uline{Takeaway 4: While edge and consumer GPUs struggle with larger models, datacenter GPUs can still support real-time inference for VLA models that are more than one order of magnitude larger.}}

\Cref{tab:pi0_model_scaling} summarizes inference performance across different model scales.
B100 sustains 9.6~Hz inference even for the largest 81B model variant (30$\times$ larger than $\pi_0$), demonstrating that modern datacenter GPUs can accommodate substantially larger VLA models under real-time constraints.
In contrast, RTX~4090 runs out of memory for $\pi_0$-XL (16.7B), and Jetson Thor struggles to deliver real-time performance even with sufficient memory capacity, achieving only 2.1~Hz inference frequency on $\pi_0$-XL.

\subsection{Long-Context VLA Inference}
\label{sec:eval:longcontext}

While $\pi_0$ model ipredicts actions solely based on the current observation, this a memory-less design is insufficient for long-horizon tasks that require reasoning over temporal context~\cite{jang2025contextvla, shi2025memoryvla, team2024octo, wang2025karma}.
In this section, we adapt $\pi_0$ to a stateful setting by enabling it to incorporate past visual states into the VLM KV cache.
At each new timestep, the latest visual inputs (three camera images, corresponding to 768 vision tokens) attend over the accumulated KV cache of the VLM, and the action prediction is conditioned on this long context.

\textit{\uline{Takeaway 5: Datacenter GPUs can support real-time long-context VLA inference with up to 1K past timesteps, while edge and consumer GPUs are limited to roughly 100 steps.}}

\Cref{tab:pi0_long_context} reports inference performance and memory consumption of long-context VLA with up to 10K past timesteps.
B100 sustains 11.7~Hz inference with 1K past timesteps, whereas performance drops to 1.2~Hz at 10K steps, which no longer meets real-time requirements.
For Jetson Thor and RTX~4090, real-time performance is only achievable when the context length is limited to roughly 100 timesteps~(around 8~Hz).

\begin{table}[t]
    \centering
    \setlength\dashlinedash{0.2pt}
    \setlength\dashlinegap{1.5pt}
    \caption{Inference performance and memory consumption of long-context VLA models.}
    \label{tab:pi0_long_context}
    \vspace{0.5em}
    \scalebox{0.9}{
\begin{tabular}{@{} L{4em} R{6em} R{6em} R{7em} R{7em} R{7em} @{}}
\toprule
\textbf{Timesteps} & \multicolumn{1}{c}{\textbf{Total Memory}} & \multicolumn{1}{c}{\textbf{KV Cache Size}} & \multicolumn{1}{c}{\textbf{Jetson Thor}} & \multicolumn{1}{c}{\textbf{RTX 4090}} & \multicolumn{1}{c}{\textbf{B100}} \\
\midrule
1 & 5.1 GB & 0.01 GB & \cellcolor{lightgray}52.6 ms (19.0 Hz) & \cellcolor{lightgray}31.1 ms (32.2 Hz) & \cellcolor{lightgray}3.2 ms (314.4 Hz) \\
\hdashline
10 & 5.3 GB & 0.13 GB & \cellcolor{lightgray}58.4 ms (17.1 Hz) & \cellcolor{lightgray}39.0 ms (25.7 Hz) & \cellcolor{lightgray}3.9 ms (254.6 Hz) \\
\hdashline
100 & 6.4 GB & 1.3 GB & \cellcolor{lightgray}122.9 ms (8.1 Hz) & \cellcolor{lightgray}117.3 ms (8.5 Hz) & \cellcolor{lightgray}11.3 ms (88.4 Hz) \\
\hdashline
1000 & 18.3 GB & 13.2 GB & \cellcolor{lightgray}768.3 ms (1.3 Hz) & \cellcolor{lightgray}900.6 ms (1.1 Hz) & \cellcolor{lightgray}85.2 ms (11.7 Hz) \\
\hdashline
10000 & 137.0 GB & 131.8 GB & N/A & N/A & \cellcolor{lightgray}823.7 ms (1.2 Hz) \\
\bottomrule
\end{tabular}
}
\end{table}

\subsection{Impact of Denoising Steps and Action Chunk Size}
\label{sec:eval:diffusion-steps}

Given a diffusion-based action expert model, two key parameters influence inference performance: (1) the number of denoising steps, where each step incurs a forward pass, and (2) the action chunk size, i.e., the number of predicted actions.
To this end, we vary the number of diffusion steps of $\pi_0$ from 1 to 50 (default: 10) and the action chunk size from 5 to 250 (default: 50), each spanning a $50\times$ range.
For brevity, we present results on B100, but the observed trends below are consistent across all evaluated GPUs.

\textit{\uline{Takeaway 6: Denoising steps have a significant impact on both action expert latency and end-to-end VLA latency, whereas action chunk size has a negligible effect.}}

\begin{figure}[t]
  \centering
  \begin{subfigure}{0.38\linewidth}
    \includegraphics[width=\linewidth]
    {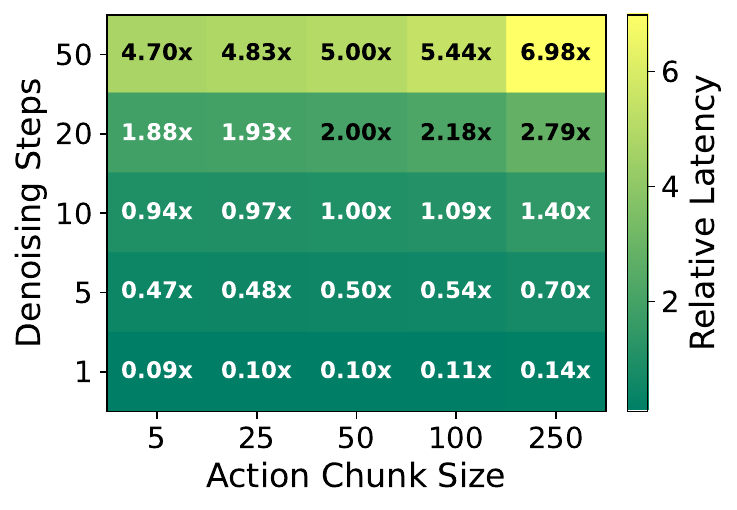}
    \caption{Action Expert Latency.}
    \label{fig:eval:denoise_steps_action_lengths:diffusion_relative}
  \end{subfigure}
  \begin{subfigure}{0.38\linewidth}
    \includegraphics[width=\linewidth]
    {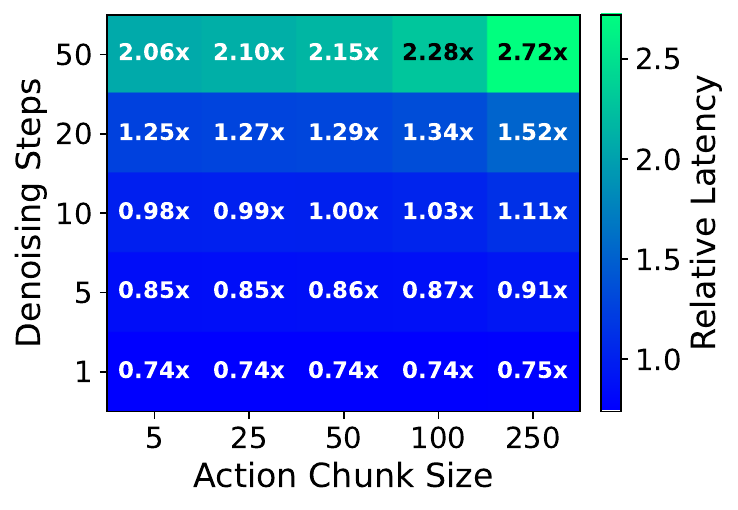}
    \caption{VLA Total Latency.}
    \label{fig:eval:denoise_steps_action_lengths:e2e_relative}
  \end{subfigure}
  \begin{subfigure}{0.22\linewidth}
    \includegraphics[width=\linewidth]
    {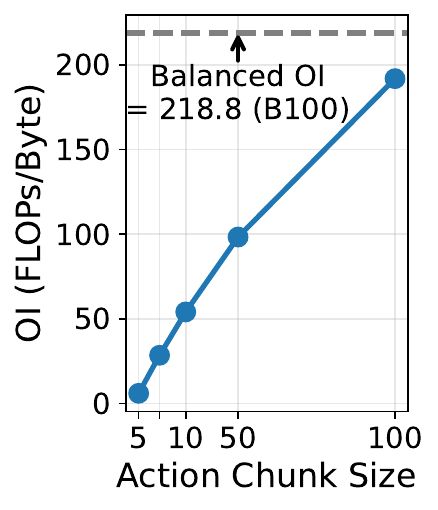}
    \caption{Action Expert OI.}
    \label{fig:eval:denoise_steps_action_lengths:oi}
  \end{subfigure}
  \caption{The impact of denoising steps and action chunk sizes to inference performance on B100 GPU.}
  \label{fig:eval:denoise_steps_action_lengths}
\end{figure}

\Cref{fig:eval:denoise_steps_action_lengths:diffusion_relative} and \Cref{fig:eval:denoise_steps_action_lengths:e2e_relative} report the action expert latency and end-to-end VLA inference latency, respectively.
On the one hand, action prediction latency scales linearly with the number of diffusion steps and thus has a substantial impact on overall VLA latency.
For example, with the default action chunk size of 50, increasing the number of diffusion steps from 10 to 50 leads to a proportional increase in action prediction latency ($5\times$) and a $2.15\times$ increase in overall VLA latency.
On the other hand, action chunk size has only a marginal effect on both action-expert latency and end-to-end VLA inference latency.
With the default setting of 10 denoising steps, increasing the action chunk size from 50 to 250 ($5\times$) increases action prediction latency by only 40\%, resulting in just an 11\% increase in end-to-end VLA latency.
This is because action prediction is typically memory-bound (\Cref{fig:eval:denoise_steps_action_lengths:oi}): performance is limited by loading model parameters and KV cache from memory, and the additional computation given more action tokens has little effect on overall latency.

\subsection{Diffusion-Based vs.\ Autoregressive Action Prediction}
\label{sec:eval:diffusion-vs-ar}

Diffusion-based and autoregressive action prediction are two dominant paradigms in recent VLA models.
Autoregressive action decoder typically uses the same transformer to process vision and language inputs and to generate actions~\cite{kim2024openvla, team2024octo}.
Accordingly, we adapt $\pi_0$ to an autoregressive variant that uses the VLM backbone directly for action prediction.
In contrast, diffusion-based VLAs usually employ a separate action expert that is significantly smaller than the VLM (e.g., $6.7\times$ smaller in $\pi_0$)~\cite{black2024pi0, bjorck2025gr00t}.
For fairness, we additionally evaluate a diffusion-based variant with an action expert that matches the VLM size, referred to as \textit{Diffusion-Large}.
Classic autoregressive VLAs generate one action dimension at a time, which results in high inference latency as it requires many sequential prediction steps (e.g., 700 steps in our case with a 14-DoF action space and an action chunk size of 50).
Thus, we also evaluate a faster autoregressive variant with parallel decoding~\cite{kim2025fine}, which predicts all actions in a single inference, denoted as \textit{Autoregressive-Parallel}.

\textit{\uline{Takeaway 7: With action chunking, diffusion-based VLA inference is one to two orders of magnitude faster than the vanilla autoregressive VLA.}}

\begin{figure}[t]
  \centering
  \begin{subfigure}{0.48\linewidth}
    \includegraphics[width=\linewidth]
    {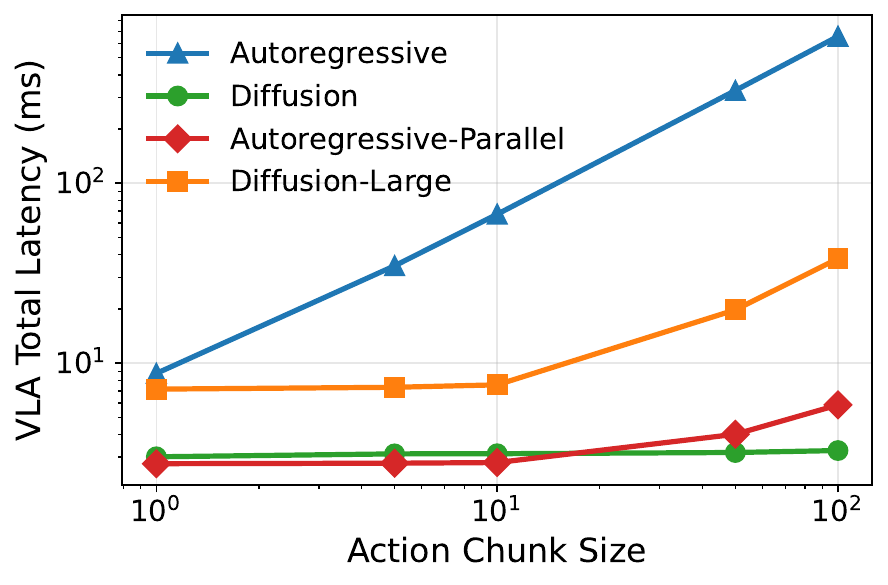}
    \caption{Varying action chunk sizes.}
    \label{fig:eval:autoregressive_diffusion:chunk_size}
  \end{subfigure}
  \begin{subfigure}{0.48\linewidth}
    \includegraphics[width=\linewidth]
    {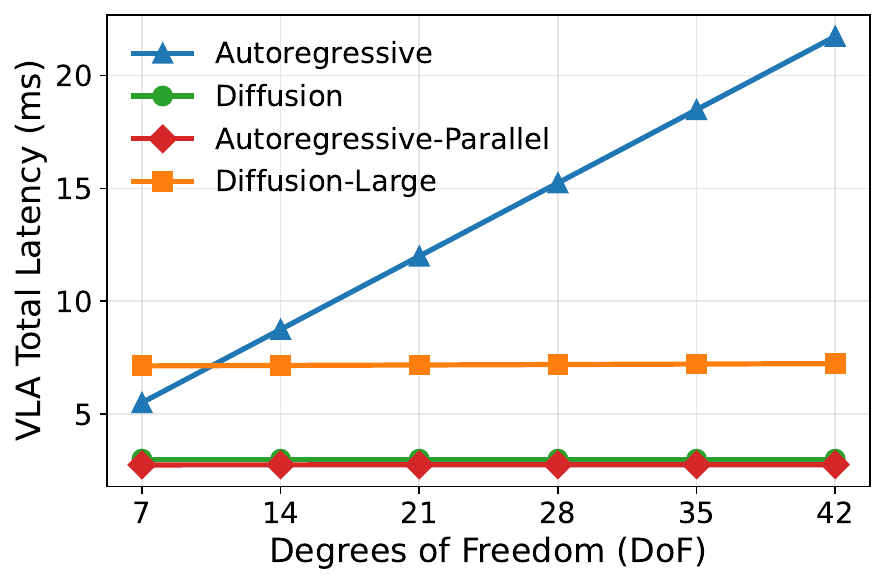}
    \caption{Varying DoF (no action chunking).}
    \label{fig:eval:autoregressive_diffusion:dof}
  \end{subfigure}
  \caption{Diffusion vs autoregressive VLA inference performance on B100 GPU.}
  \label{fig:eval:autoregressive_diffusion}
\end{figure}

\Cref{fig:eval:autoregressive_diffusion:chunk_size} compares inference latency across different architectures on the B100 GPU.
Across all action chunk sizes, diffusion-based models (both the standard and large variants) consistently outperform the vanilla autoregressive VLA.
With the default chunk size of 50, the standard diffusion model achieves an inference latency of 3.2~ms, which is $102.4\times$ faster than the classic autoregressive model (327.6~ms).

\textit{\uline{Takeaway 8: Autoregressive VLAs are competitive only when generating a small number of action tokens or when parallel decoding is enabled.}}

To further analyze scenarios where autoregressive VLAs can be efficient, we evaluate inference performance without action chunking across common action dimensionalities, ranging from 7~DoF for a single robot arm~\cite{black2024pi0, zitkovich2023rt} to over 40~DoF for two dexterous hands~\cite{wen2025dexterous, christoph2025orca}.
As shown in \Cref{fig:eval:autoregressive_diffusion:dof}, the autoregressive model can slightly outperform the large diffusion-based model when the number of generated action tokens is small (e.g., 7), although the standard-size diffusion model remains faster.
Another scenario in which autoregressive inference becomes competitive is when parallel decoding is employed.
As shown in \Cref{fig:eval:autoregressive_diffusion:chunk_size},  parallel decoding outperforms the standard diffusion model for action chunk sizes up to 10.
However, for larger chunk sizes such as 50, the latency of parallel decoding increases substantially as the workload transitions from memory-bound to compute-bound (OI increases from 135.9 at chunk size 10 to 477.7 at chunk size 50, exceeding the B100 balance OI of 218.8).
In contrast, the diffusion-based action expert remains memory-bound (OI~=~54.0 at chunk size of 50), leading to more stable inference performance across chunk sizes.

\subsection{On-Device vs. Server-Side Inference}
\label{sec:eval:device_vs_server}

We evaluate three classes of VLA inference systems with different GPU and network configurations.
First, \emph{on-device inference}, where inference is executed directly on an edge GPU integrated into the robot (e.g., Jetson Thor), as demonstrated by systems such as Figure AI’s Helix~\cite{figure2025helix}.
Second, \emph{edge-server inference}, where inference is performed on a server located close to the robot~\cite{black2024pi0, bjorck2025gr00t, huang2025dadu}.
In this setting, communication between the robot and the server may use wired networks (Ethernet) for fixed-base robots or wireless networks (WiFi or cellular networks) for mobile robots (e.g., wheeled or humanoid robots), while the server-side accelerator may range from consumer-grade GPUs (e.g., RTX~4090) to datacenter-class GPUs (e.g., B100).
Third, \emph{cloud-server inference}, where inference runs on high-end datacenter GPUs.
In this case, the robot first communicates with a nearby gateway server via a wired or wireless connection, which then forwards inference requests to the cloud, incurring two stages of communication latency.
Note that network latency to cloud servers can vary substantially, depending on factors such as physical distance and routing topology~\cite{mok2021measuring, sfiligoi2020characterizing}; thus, we consider two cloud network configurations with different performance.
We summarize detailed network performance parameters in \Cref{tab:network_configs}. 

\textit{\uline{Takeaway 9: Server-side inference, even with only consumer GPUs, significantly outperforms on-device inference in most scenarios, except under extremely poor network conditions.}}

\begin{figure}[t]
  \centering
  \includegraphics[width=\linewidth]{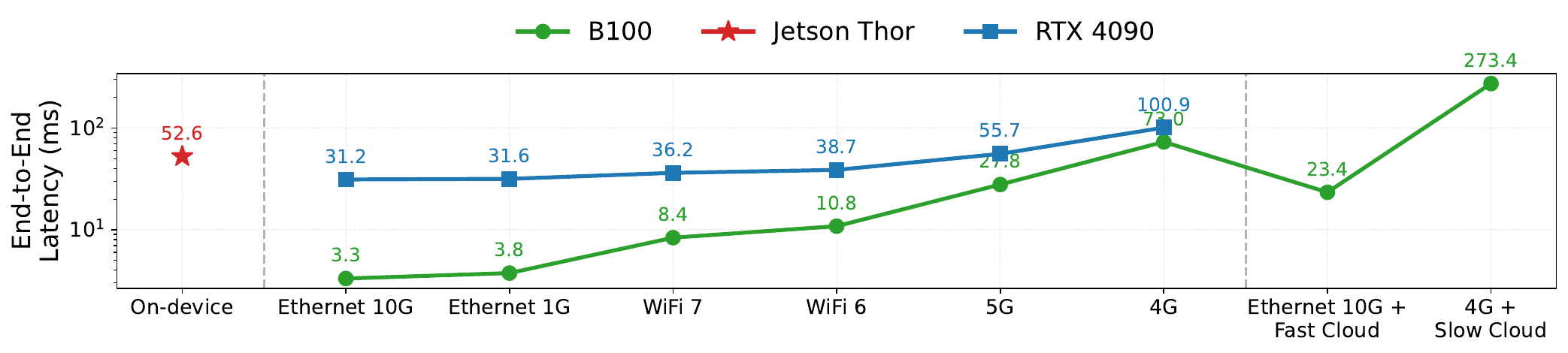}
  \vspace{-1em}
  \caption{Inference performance on device, on edge servers, and on cloud servers.}
  \label{fig:eval:device_vs_server}
\end{figure}

As shown in \Cref{fig:eval:device_vs_server}, even when using a consumer GPU (RTX~4090) connected via WiFi, server-side inference achieves lower end-to-end latency than on-device inference on Jetson Thor.
With a more powerful B100 GPU, inference remains faster than on-device execution even when deployed on an edge server with only cellular (5G) connectivity or in a cloud instance with fast network.
On-device inference becomes preferable only when network conditions are extremely constrained, such as (i) slow cellular connections (4G or below), or (ii) cloud deployments where the datacenter is distant from the robot.

\subsection{Device-Server Collaborative Inference}
\label{sec:eval:device_server_collab}

Some robots already have an on-board GPU --- so a natural idea is to split the VLA workload between server and device to (1) reduce server workloads and to (2) improve performance over device-only deployments.
Since the action expert model is usually several times smaller than the VLM backbone~\cite{black2024pi0, shukor2025smolvla, wen2025tinyvla}, a natural idea is to run VLM inference (including the vision encoder) on server (B100) and run action expert inference on device (Jetson Thor).
In comparison to either device-only or server-only solutions, here, device-server collaboration involves an extra communication step, where the KV cache of the VLM has to be downloaded to the device GPU before action prediction begins.

\textit{\uline{Takeaway 10. Device–server collaboration is often slower than device-only inference and always slower than server-side inference, making this solution generally unattractive in practice.}}

\begin{figure}[t]
  \centering
  \includegraphics[width=\linewidth]{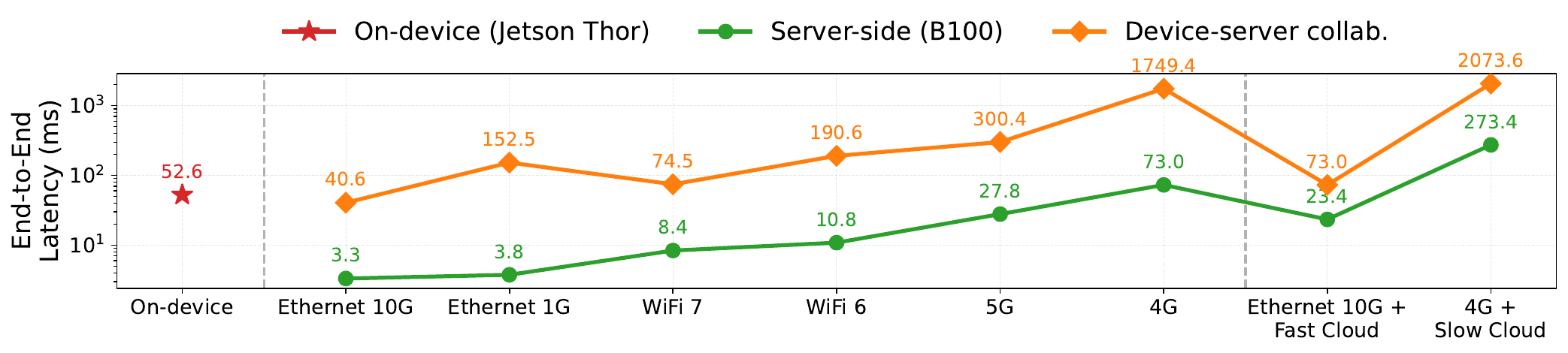}
  \vspace{-1em}
  \caption{Device-server collaborative inference versus server-only and device-only solutions.}

  \label{fig:eval:device_server_collab}
\end{figure}

As shown in \Cref{fig:eval:device_server_collab}, collaborative inference is always slower compared to server-only inference --- which is not surprising as now the action expert runs on a less powerful device.
What we found interesting is that it is even slower than on-device inference in most cases, except with a fast wired network (Ethernet 10G) --- this is because of the KV cache download process from the server to the device, which can be very slow without a fast network (12.4, 43.7, and 257.7 ms for Ethernet 10G, WiFI 7, and 5G networks, respectively).
However, we argue that such scenarios are rare in practice: robots equipped with on-device GPUs are typically mobile platforms that relies on wireless connectivity, in which case using the on-device GPU alone (rather than device-server collaboration) is the more performant choice.

\begin{table}[t]
    \centering
    \setlength\dashlinedash{0.2pt}
    \setlength\dashlinegap{1.5pt}
    \caption{Network configuration specifications.}
    \label{tab:network_configs}
    \vspace{0.5em}
    \scalebox{0.72}{
\begin{tabular}{@{} L{6em} M{6em} M{6em} M{4em} M{4em} M{4em} M{4em} M{5em} M{5em} @{}}
\toprule
\textbf{Metric} & \multicolumn{1}{c}{\textbf{Ethernet 1G}} & \multicolumn{1}{c}{\textbf{Ethernet 10G}} & \multicolumn{1}{c}{\textbf{WiFi 6}} & \multicolumn{1}{c}{\textbf{WiFi 7}} & \multicolumn{1}{c}{\textbf{4G}} & \multicolumn{1}{c}{\textbf{5G}} & \multicolumn{1}{c}{\textbf{Slow Cloud}} & \multicolumn{1}{c}{\textbf{Fast Cloud}} \\
\midrule
Upload BW & 1 Gbps & 10 Gbps & 560 Mbps & 2 Gbps & 19 Mbps & 80 Mbps & 1 Gbps & 10 Gbps \\
\hdashline
Download BW & 1 Gbps & 10 Gbps & 800 Mbps & 3 Gbps & 75 Mbps & 500 Mbps & 1 Gbps & 10 Gbps \\
\hdashline
Base Latency & 0.10 ms & 0.05 ms & 3.50 ms & 2.50 ms & 25.00 ms & 10.00 ms & 100.00 ms & 10.00 ms \\
\bottomrule
\end{tabular}
}
\end{table}

\subsection{Asynchronous Inference}
\label{sec:eval:async}

With asynchronous inference, the model predicts actions based on stale observations rather than the latest state, allowing model inference and robot action execution to be partially overlapped.
While this form of asynchrony does not increase the maximum inference throughput for on-device inference without network latency, it can benefit server-side inference by allowing network transmission and GPU computation to proceed concurrently.
Thus, the asynchronous inference throughput is bounded by the minimum of GPU inference throughput and network transmission throughput.

\textit{\uline{Takeaway 11: Asynchrony between robot execution and inference can significantly improve system throughput for server-side inference, especially under slow wireless network connections.}}

\Cref{tab:async_inference_robot} reports the server-side inference throughput under different network configurations.
With fast wired networks (e.g., 1~GbE and 10~GbE Ethernet), synchronous and asynchronous inference achieve similar throughput.
In contrast, under slower wireless networks (WiFi~7, 5G, and 4G), asynchronous inference improves throughput by 2.63$\sim$5.99$\times$.
With WiFi~7, inference remains GPU-bound, and thus achieves the same throughput to wired networks (314.4Hz).
For 5G and 4G, the bottleneck shifts to network transmission, resulting in lower asynchronous throughput.
Note that while asynchronous inference improves throughput, it does not reduce end-to-end latency; increased staleness may degrade action quality, which warrants further investigation from the perspective of control stability and task success rate.

\begin{table}[t]
    \centering
    \setlength\dashlinedash{0.2pt}
    \setlength\dashlinegap{1.5pt}
    \caption{Inference frequency of synchronous and asynchronous systems.}
    \label{tab:async_inference_robot}
    \vspace{0.5em}
    \scalebox{0.85}{
\begin{tabular}{@{} L{5em} L{8em} R{4em} R{6em} R{6em} R{4em} @{}}
\toprule
\textbf{Hardware} & \textbf{Network} & \textbf{Latency} & \textbf{Freq. (Sync)} & \textbf{Freq. (Async)} & \textbf{Speedup} \\
\midrule
B100 & Ethernet 10G & 3.3 ms & 301.4 Hz & 314.4 Hz & \cellcolor{lightgray} 1.04$\times$ \\
\hdashline
B100 & Ethernet 1G & 3.8 ms & 266.5 Hz & 314.4 Hz & \cellcolor{lightgray} 1.18$\times$ \\
\hdashline
B100 & WiFi 7 & 8.4 ms & 119.7 Hz & 314.4 Hz & \cellcolor{lightgray} 2.63$\times$ \\
\hdashline
B100 & 5G & 27.8 ms & 35.9 Hz & 215.3 Hz & \cellcolor{lightgray} 5.99$\times$ \\
\hdashline
B100 & 4G & 73.0 ms & 13.7 Hz & 50.5 Hz & \cellcolor{lightgray} 3.68$\times$ \\
\midrule
B100 & Wired + Fast Cloud & 23.4 ms & 42.8 Hz & 314.4 Hz & \cellcolor{lightgray} 7.34$\times$ \\
\hdashline
B100 & 4G + Slow Cloud & 273.4 ms & 3.7 Hz & 50.5 Hz & \cellcolor{lightgray} 13.79$\times$ \\
\bottomrule
\end{tabular}
}
\end{table}

\subsection{Dual-system VLA Pipelines}
\label{sec:eval:dual_system}

Recent work proposes a \textit{System~1 + System~2} paradigm for action generation, where a slower System~2 (the VLM) responsible for high-level reasoning operates at a lower frequency (e.g., 5--10~Hz), while a faster System~1 (the action model) reacts to the environment at a higher frequency using the most recent visual inputs~\cite{figure2025helix, zhang2024hirt, song2025hume}.
The two systems run asynchronously: the action expert conditions its predictions on the VLM’s KV cache, which is updated at a lower frequency by System~2.
While this design is conceptually appealing, we are not aware of a widely adopted, open-source diffusion-style implementation of a dual-system VLA.
Therefore, we make the following approximations in our evaluation:
(1) System~1 latency consists of image upload, vision encoding, diffusion-based action prediction, and action download, where the cost of integrating vision features into the action expert is assumed to be negligible; and
(2) System~2 latency equals VLM inference, which incorporates the visual encoding of the most recently uploaded image.

\textit{\uline{Takeaway 12: Asynchronous inference between System 1 and System 2 can improve action prediction performance, with performance gains strongly dependent on hardware capability and network latency.}}

\Cref{tab:pi0_system_1_2} reports performance across different system configurations and System~2 frequency caps (5~Hz and 10~Hz).
On Jetson Thor, the improvement is moderate (1.46$\times$ at a 5~Hz cap and 1.30$\times$ at a 10~Hz cap), since the asynchronous frequency cap is comparable to the synchronous VLM frequency of 19~Hz.
In contrast, on B100 with a fast 10G Ethernet connection, the speedup is substantial (2.24$\times$ at a 10~Hz cap).
In this case, asynchronous execution significantly reduces the effective VLM invocation rate from 301.4~Hz to 10~Hz, freeing compute resources that can be reallocated to action prediction.
However, under slower network conditions (e.g., 5G), the benefit diminishes, yielding only a 1.05$\times$ speedup at a 10~Hz cap.
This is because network latency substantially increases System~1 latency --- from 1.5~ms with Ethernet to 26.0~ms with 5G --- thereby limiting the achievable performance regardless of the inference hardware capability.

\begin{table}[t]
\centering
\setlength\dashlinedash{0.2pt}
\setlength\dashlinegap{1.5pt}
\caption{Performance gains by using dual-system inference.}
\label{tab:pi0_system_1_2}
\vspace{0.3em}
\scalebox{0.76}{
\begin{tabular}{@{}
L{5em} L{5em}
R{3em} R{3em} R{5em}
M{0em}
R{5.5em} R{3.5em}
M{0em}
R{5.5em} R{3.5em}
@{}}
\toprule
\multirow{2}{*}{\textbf{Hardware}}
& \multirow{2}{*}{\textbf{Network}}
& \multirow{2}{*}{\textbf{S1 Lat.}}
& \multirow{2}{*}{\textbf{S2 Lat.}}
& \multirow{2}{*}{\textbf{Freq. (Sync)}}
& 
& \multicolumn{2}{c}{\textbf{S2 Cap = 5 Hz}}
& 
& \multicolumn{2}{c}{\textbf{S2 Cap = 10 Hz}} \\
\cmidrule(lr){7-8}
\cmidrule(lr){10-11}
& & & & &
& \textbf{Freq. (Async)} & \textbf{Speedup}
&
& \textbf{Freq. (Async)} & \textbf{Speedup} \\
\midrule
Jetson Thor
& On-device
& 32.3 ms
& 20.3 ms
& 19.0 Hz
&& 27.8 Hz & \cellcolor{lightgray} 1.46$\times$
&& 24.7 Hz & \cellcolor{lightgray} 1.30$\times$ \\
\hdashline
B100
& Ethernet 10G
& 1.5 ms
& 1.9 ms
& 301.4 Hz
&& 682.4 Hz & \cellcolor{lightgray} 2.26$\times$
&& 676.0 Hz & \cellcolor{lightgray} 2.24$\times$ \\
\hdashline
B100
& WiFi~7
& 6.5 ms
& 1.9 ms
& 119.7 Hz
&& 152.6 Hz & \cellcolor{lightgray} 1.28$\times$
&& 151.2 Hz & \cellcolor{lightgray} 1.26$\times$ \\
\hdashline
B100
& 5G
& 26.0 ms
& 1.9 ms
& 35.9 Hz
&& 38.2 Hz & \cellcolor{lightgray} 1.06$\times$
&& 37.8 Hz & \cellcolor{lightgray} 1.05$\times$ \\
\bottomrule
\end{tabular}
}
\end{table}

\subsection{Supporting High-Performance VLA Inference up to 100~Hz}
\label{sec:eval:frequency}

In this section, we analyze how 10~Hz and 100~Hz performance targets~(\S\ref{sec:background}) can be achieved with the $\pi_0$ model across on-device, edge-server, and cloud-server inference systems.
We also discuss what algorithm-level adjustments may be required when the target performance cannot be met by those systems.

\textit{\uline{Takeaway 13: For on-device inference, the most advanced edge GPUs (Jetson Thor) can already achieve 10~Hz inference for $\pi_0$, but reaching 100~Hz requires model-level adjustments.}}

\Cref{tab:pi0_base_performance} shows that Jetson Thor already achieves 19~Hz inference throughput for $\pi_0$, exceeding the 10~Hz target.
However, achieving 100~Hz would require roughly a $5\times$ improvement.
This gap have to be closed through model-level optimizations, such as reducing model size (\S\ref{sec:eval:scaling}), decreasing the number of diffusion steps (\S\ref{sec:eval:diffusion-steps}), or using lower-precision quantization~\cite{kim2024openvla, wang2025bitvla}.

\textit{\uline{Takeaway 14: For edge-server inference, 10~Hz is achievable with consumer GPUs and wireless networks, while 100~Hz requires datacenter GPUs and faster networks.}}

\Cref{fig:eval:device_vs_server} shows that an RTX~4090 can achieve 10~Hz inference even with a slow 4G network.
However, achieving sub-10~ms latency (100~Hz) requires either a more powerful accelerator such as B100 or the aforementioned model-level optimizations.
For B100, reaching 100~Hz further depends on network performance, requiring either wired Ethernet or high-quality wireless connectivity (e.g., WiFi~7).

\textit{\uline{Takeaway 15: For cloud-server inference, 10~Hz is feasible with good networking, while achieving 100~Hz generally requires asynchronous inference.}}

As shown in \Cref{tab:async_inference_robot}, B100 achieves only 42.8~Hz under synchronous cloud inference even with a fast network.
In this regime, network latency alone (exceeding 10~ms per upload or download) prevents achieving 100~Hz, making computation reduction insufficient and rendering asynchronous inference necessary for high-frequency operation.
With a fast network (WiFi~7 or better), asynchronous execution cab achieve a throughput of 314.4~Hz.
Even under poor network conditions where synchronous inference becomes unacceptable (3.7~Hz), asynchronous inference can still restore acceptable performance (50.5~Hz).

\section{Conclusion and Future Work}
\label{sec:conclusion}

We present the first comprehensive study of VLA inference performance.
Using \ours, an analytical performance modeling tool that we develop, we systematically explore a wide range of (1) model configurations, including model size, context length, architectural choices, and synchronous versus asynchronous execution, and (2) system configurations spanning different hardware platforms, inference placements, and network conditions.
From the performance study, we distill 15 key takeaways that provide practical guidance for the design of future VLA models and inference systems.

While this work represents an important step toward understanding and building next-generation VLA systems, we view it as only a starting point.
First, our study focuses primarily on VLA models for manipulation tasks and does not consider other embodied AI domains such as autonomous driving, quadrupeds, or drones.
These settings often involve different system constraints (e.g., stronger emphasis on on-device execution) and additional model components (e.g., SLAM and specialized control modules), which are beyond the scope of this work.
Second, robotic systems are complex end-to-end pipelines that go beyond model inference alone.
In this work, we do not account for robot execution latency or sensor latency (e.g., cameras), as these factors vary widely across platforms.
A more comprehensive performance analysis that integrates inference, sensing, and actuation would provide deeper insights into end-to-end robotic system behavior.
We leave these directions to future work.

\bibliographystyle{unsrt}
\bibliography{refs}

@article{wen2025tinyvla,
  title={Tinyvla: Towards fast, data-efficient vision-language-action models for robotic manipulation},
  author={Wen, Junjie and Zhu, Yichen and Li, Jinming and Zhu, Minjie and Tang, Zhibin and Wu, Kun and Xu, Zhiyuan and Liu, Ning and Cheng, Ran and Shen, Chaomin and others},
  journal={IEEE Robotics and Automation Letters},
  year={2025},
  publisher={IEEE}
}

@article{shukor2025smolvla,
  title={Smolvla: A vision-language-action model for affordable and efficient robotics},
  author={Shukor, Mustafa and Aubakirova, Dana and Capuano, Francesco and Kooijmans, Pepijn and Palma, Steven and Zouitine, Adil and Aractingi, Michel and Pascal, Caroline and Russi, Martino and Marafioti, Andres and others},
  journal={arXiv preprint arXiv:2506.01844},
  year={2025}
}

@article{lin2025evo,
  title={Evo-1: Lightweight vision-language-action model with preserved semantic alignment},
  author={Lin, Tao and Zhong, Yilei and Du, Yuxin and Zhang, Jingjing and Liu, Jiting and Chen, Yinxinyu and Gu, Encheng and Liu, Ziyan and Cai, Hongyi and Zou, Yanwen and others},
  journal={arXiv preprint arXiv:2511.04555},
  year={2025}
}

@misc{figure2025helix,
  title        = {Helix: A Vision-Language-Action Model for Generalist Humanoid Control},
  author       = {{Figure AI}},
  year         = {2025},
  howpublished = {\url{https://www.figure.ai/news/helix}},
}

@article{zhang2024hirt,
  title={Hirt: Enhancing robotic control with hierarchical robot transformers},
  author={Zhang, Jianke and Guo, Yanjiang and Chen, Xiaoyu and Wang, Yen-Jen and Hu, Yucheng and Shi, Chengming and Chen, Jianyu},
  journal={arXiv preprint arXiv:2410.05273},
  year={2024}
}

@article{song2025hume,
  title={Hume: Introducing System-2 Thinking in Visual-Language-Action Model},
  author={Song, Haoming and Qu, Delin and Yao, Yuanqi and Chen, Qizhi and Lv, Qi and Tang, Yiwen and Shi, Modi and Ren, Guanghui and Yao, Maoqing and Zhao, Bin and others},
  journal={arXiv preprint arXiv:2505.21432},
  year={2025}
}

@article{jang2025contextvla,
  title={ContextVLA: Vision-Language-Action Model with Amortized Multi-Frame Context},
  author={Jang, Huiwon and Yu, Sihyun and Kwon, Heeseung and Jeon, Hojin and Seo, Younggyo and Shin, Jinwoo},
  journal={arXiv preprint arXiv:2510.04246},
  year={2025}
}

@article{shi2025memoryvla,
  title={Memoryvla: Perceptual-cognitive memory in vision-language-action models for robotic manipulation},
  author={Shi, Hao and Xie, Bin and Liu, Yingfei and Sun, Lin and Liu, Fengrong and Wang, Tiancai and Zhou, Erjin and Fan, Haoqiang and Zhang, Xiangyu and Huang, Gao},
  journal={arXiv preprint arXiv:2508.19236},
  year={2025}
}

@article{team2024octo,
  title={Octo: An open-source generalist robot policy},
  author={Team, Octo Model and Ghosh, Dibya and Walke, Homer and Pertsch, Karl and Black, Kevin and Mees, Oier and Dasari, Sudeep and Hejna, Joey and Kreiman, Tobias and Xu, Charles and others},
  journal={arXiv preprint arXiv:2405.12213},
  year={2024}
}

@article{black2025real,
  title={Real-Time Execution of Action Chunking Flow Policies},
  author={Black, Kevin and Galliker, Manuel Y and Levine, Sergey},
  journal={arXiv preprint arXiv:2506.07339},
  year={2025}
}

@article{black2025training,
  title={Training-Time Action Conditioning for Efficient Real-Time Chunking},
  author={Black, Kevin and Ren, Allen Z and Equi, Michael and Levine, Sergey},
  journal={arXiv preprint arXiv:2512.05964},
  year={2025}
}

@article{sendai2025leave,
  title={Leave no observation behind: Real-time correction for vla action chunks},
  author={Sendai, Kohei and Alvarez, Maxime and Matsushima, Tatsuya and Matsuo, Yutaka and Iwasawa, Yusuke},
  journal={arXiv preprint arXiv:2509.23224},
  year={2025}
}

@article{tang2025vlash,
  title={VLASH: Real-Time VLAs via Future-State-Aware Asynchronous Inference},
  author={Tang, Jiaming and Sun, Yufei and Zhao, Yilong and Yang, Shang and Lin, Yujun and Zhang, Zhuoyang and Hou, James and Lu, Yao and Liu, Zhijian and Han, Song},
  journal={arXiv preprint arXiv:2512.01031},
  year={2025}
}

@article{brohan2022rt,
  title={Rt-1: Robotics transformer for real-world control at scale},
  author={Brohan, Anthony and Brown, Noah and Carbajal, Justice and Chebotar, Yevgen and Dabis, Joseph and Finn, Chelsea and Gopalakrishnan, Keerthana and Hausman, Karol and Herzog, Alex and Hsu, Jasmine and others},
  journal={arXiv preprint arXiv:2212.06817},
  year={2022}
}

@inproceedings{zitkovich2023rt,
  title={Rt-2: Vision-language-action models transfer web knowledge to robotic control},
  author={Zitkovich, Brianna and Yu, Tianhe and Xu, Sichun and Xu, Peng and Xiao, Ted and Xia, Fei and Wu, Jialin and Wohlhart, Paul and Welker, Stefan and Wahid, Ayzaan and others},
  booktitle={Conference on Robot Learning},
  pages={2165--2183},
  year={2023},
  organization={PMLR}
}

@article{gu2023rt,
  title={Rt-trajectory: Robotic task generalization via hindsight trajectory sketches},
  author={Gu, Jiayuan and Kirmani, Sean and Wohlhart, Paul and Lu, Yao and Arenas, Montserrat Gonzalez and Rao, Kanishka and Yu, Wenhao and Fu, Chuyuan and Gopalakrishnan, Keerthana and Xu, Zhuo and others},
  journal={arXiv preprint arXiv:2311.01977},
  year={2023}
}

@article{black2024pi0,
  title={$\pi$0: A visionlanguage-action flow model for general robot control, 2024a},
  author={Black, Kevin and Brown, Noah and Driess, Danny and Esmail, Adnan and Equi, Michael and Finn, Chelsea and Fusai, Niccolo and Groom, Lachy and Hausman, Karol and Ichter, Brian and others},
  journal={URL https://arxiv. org/abs/2410.24164},
  year={2024}
}

@article{intelligence2504pi0,
  title={$\pi$0.5: A vision-language-action model with open-world generalization. arXiv 2025},
  author={Intelligence, P and Black, K and Brown, N and Darpinian, J and Dhabalia, K and Driess, D and Esmail, A and Equi, M and Finn, C and Fusai, N and others},
  journal={arXiv preprint arXiv:2504.16054}
}

@article{amin2025pi,
  title={$\pi$0.6: a VLA That Learns From Experience},
  author={Amin, Ali and Aniceto, Raichelle and Balakrishna, Ashwin and Black, Kevin and Conley, Ken and Connors, Grace and Darpinian, James and Dhabalia, Karan and DiCarlo, Jared and Driess, Danny and others},
  journal={arXiv preprint arXiv:2511.14759},
  year={2025}
}

@article{team2025gemini,
  title={Gemini robotics: Bringing ai into the physical world},
  author={Team, Gemini Robotics and Abeyruwan, Saminda and Ainslie, Joshua and Alayrac, Jean-Baptiste and Arenas, Montserrat Gonzalez and Armstrong, Travis and Balakrishna, Ashwin and Baruch, Robert and Bauza, Maria and Blokzijl, Michiel and others},
  journal={arXiv preprint arXiv:2503.20020},
  year={2025}
}

@article{bjorck2025gr00t,
  title={Gr00t n1: An open foundation model for generalist humanoid robots},
  author={Bjorck, Johan and Casta{\~n}eda, Fernando and Cherniadev, Nikita and Da, Xingye and Ding, Runyu and Fan, Linxi and Fang, Yu and Fox, Dieter and Hu, Fengyuan and Huang, Spencer and others},
  journal={arXiv preprint arXiv:2503.14734},
  year={2025}
}

@article{generalist2025gen0,
          author = {Generalist AI Team},
          title = {GEN-0: Embodied Foundation Models That Scale with Physical Interaction},
          journal = {Generalist AI Blog},
          year = {2025},
          note = {https://generalistai.com/blog/preview-uqlxvb-bb.html},

        }

@article{yu2025survey,
  title={A survey on efficient vision-language-action models},
  author={Yu, Zhaoshu and Wang, Bo and Zeng, Pengpeng and Zhang, Haonan and Zhang, Ji and Gao, Lianli and Song, Jingkuan and Sebe, Nicu and Shen, Heng Tao},
  journal={arXiv preprint arXiv:2510.24795},
  year={2025}
}

@article{ma2025running,
  title={Running VLAs at Real-time Speed},
  author={Ma, Yunchao and Zhou, Yizhuang and Yang, Yunhuan and Wang, Tiancai and Fan, Haoqiang},
  journal={arXiv preprint arXiv:2510.26742},
  year={2025}
}

@inproceedings{huang2025dadu,
  title={Dadu-Corki: Algorithm-Architecture Co-Design for Embodied AI-powered Robotic Manipulation},
  author={Huang, Yiyang and Hao, Yuhui and Yu, Bo and Yan, Feng and Yang, Yuxin and Min, Feng and Han, Yinhe and Ma, Lin and Liu, Shaoshan and Liu, Qiang and others},
  booktitle={Proceedings of the 52nd Annual International Symposium on Computer Architecture},
  pages={327--343},
  year={2025}
}

@article{sun2026dadu,
  title={Dadu-e: Rethinking the Role of Large Language Model in Robotic Computing Pipelines},
  author={Sun, Wenhao and Hou, Sai and Wang, Zixuan and Yu, Bo and Liu, Shaoshan and Yang, Xu and Liang, Shuai and Gan, Yiming and Han, Yinhe},
  journal={Journal of Field Robotics},
  year={2026},
  publisher={Wiley Online Library}
}

@inproceedings{wang2025karma,
  title={Karma: Augmenting embodied ai agents with long-and-short term memory systems},
  author={Wang, Zixuan and Yu, Bo and Zhao, Junzhe and Sun, Wenhao and Hou, Sai and Liang, Shuai and Hu, Xing and Han, Yinhe and Gan, Yiming},
  booktitle={2025 IEEE International Conference on Robotics and Automation (ICRA)},
  pages={1--8},
  year={2025},
  organization={IEEE}
}

@article{yue2024deer,
  title={Deer-vla: Dynamic inference of multimodal large language models for efficient robot execution},
  author={Yue, Yang and Wang, Yulin and Kang, Bingyi and Han, Yizeng and Wang, Shenzhi and Song, Shiji and Feng, Jiashi and Huang, Gao},
  journal={Advances in Neural Information Processing Systems},
  volume={37},
  pages={56619--56643},
  year={2024}
}

@article{yangdysl,
  title={DySL-VLA: Efficient Vision-Language-Action Model Inference via Dynamic-Static Layer-Skipping for Robot Manipulation},
  author={Yang, Zebin and Qi, Yijiahao and Xie, Tong and Yu, Bo and Liu, Shaoshan and Li, Meng}
}

@article{kim2025fine,
  title={Fine-tuning vision-language-action models: Optimizing speed and success},
  author={Kim, Moo Jin and Finn, Chelsea and Liang, Percy},
  journal={arXiv preprint arXiv:2502.19645},
  year={2025}
}

@article{zhao2023learning,
  title={Learning fine-grained bimanual manipulation with low-cost hardware},
  author={Zhao, Tony Z and Kumar, Vikash and Levine, Sergey and Finn, Chelsea},
  journal={arXiv preprint arXiv:2304.13705},
  year={2023}
}

@article{jing2025mixture,
  title={Mixture of Horizons in Action Chunking},
  author={Jing, Dong and Wang, Gang and Liu, Jiaqi and Tang, Weiliang and Sun, Zelong and Yao, Yunchao and Wei, Zhenyu and Liu, Yunhui and Lu, Zhiwu and Ding, Mingyu},
  journal={arXiv preprint arXiv:2511.19433},
  year={2025}
}

@article{kim2024openvla,
  title={Openvla: An open-source vision-language-action model},
  author={Kim, Moo Jin and Pertsch, Karl and Karamcheti, Siddharth and Xiao, Ted and Balakrishna, Ashwin and Nair, Suraj and Rafailov, Rafael and Foster, Ethan and Lam, Grace and Sanketi, Pannag and others},
  journal={arXiv preprint arXiv:2406.09246},
  year={2024}
}

@article{davies2025liminal,
  title={LIMINAL: Exploring The Frontiers of LLM Decode Performance},
  author={Davies, Michael and Crago, Neal and Sankaralingam, Karthikeyan and Kozyrakis, Christos},
  journal={arXiv preprint arXiv:2507.14397},
  year={2025}
}

@article{bambhaniya2024demystifying,
  title={Demystifying AI Platform Design for Distributed Inference of Next-Generation LLM models},
  author={Bambhaniya, Abhimanyu and Raj, Ritik and Jeong, Geonhwa and Kundu, Souvik and Srinivasan, Sudarshan and Subramanian, Suvinay and Elavazhagan, Midhilesh and Kumar, Madhu and Krishna, Tushar},
  journal={arXiv preprint arXiv:2406.01698},
  year={2024}
}

@article{agrawal2024vidur,
  title={Vidur: A large-scale simulation framework for llm inference},
  author={Agrawal, Amey and Kedia, Nitin and Mohan, Jayashree and Panwar, Ashish and Kwatra, Nipun and Gulavani, Bhargav S and Ramjee, Ramachandran and Tumanov, Alexey},
  journal={Proceedings of Machine Learning and Systems},
  volume={6},
  pages={351--366},
  year={2024}
}

@inproceedings{cho2024llmservingsim,
  title={Llmservingsim: A hw/sw co-simulation infrastructure for llm inference serving at scale},
  author={Cho, Jaehong and Kim, Minsu and Choi, Hyunmin and Heo, Guseul and Park, Jongse},
  booktitle={2024 IEEE International Symposium on Workload Characterization (IISWC)},
  pages={15--29},
  year={2024},
  organization={IEEE}
}

@article{yuan2024llm,
  title={Llm inference unveiled: Survey and roofline model insights},
  author={Yuan, Zhihang and Shang, Yuzhang and Zhou, Yang and Dong, Zhen and Zhou, Zhe and Xue, Chenhao and Wu, Bingzhe and Li, Zhikai and Gu, Qingyi and Lee, Yong Jae and others},
  journal={arXiv preprint arXiv:2402.16363},
  year={2024}
}

@inproceedings{jiang2025rago,
  title={Rago: Systematic performance optimization for retrieval-augmented generation serving},
  author={Jiang, Wenqi and Subramanian, Suvinay and Graves, Cat and Alonso, Gustavo and Yazdanbakhsh, Amir and Dadu, Vidushi},
  booktitle={Proceedings of the 52nd Annual International Symposium on Computer Architecture},
  pages={974--989},
  year={2025}
}

@inproceedings{patel2024splitwise,
  title={Splitwise: Efficient generative llm inference using phase splitting},
  author={Patel, Pratyush and Choukse, Esha and Zhang, Chaojie and Shah, Aashaka and Goiri, {\'I}{\~n}igo and Maleki, Saeed and Bianchini, Ricardo},
  booktitle={2024 ACM/IEEE 51st Annual International Symposium on Computer Architecture (ISCA)},
  pages={118--132},
  year={2024},
  organization={IEEE}
}

@article{wen2025dexterous,
  title={Dexterous Teleoperation of 20-DoF ByteDexter Hand via Human Motion Retargeting},
  author={Wen, Ruoshi and Zhang, Jiajun and Chen, Guangzeng and Cui, Zhongren and Du, Min and Gou, Yang and Han, Zhigang and Hu, Junkai and Huang, Liqun and Niu, Hao and others},
  journal={arXiv preprint arXiv:2507.03227},
  year={2025}
}

@article{christoph2025orca,
  title={ORCA: An Open-Source, Reliable, Cost-Effective, Anthropomorphic Robotic Hand for Uninterrupted Dexterous Task Learning},
  author={Christoph, Clemens C and Eberlein, Maximilian and Katsimalis, Filippos and Roberti, Arturo and Sympetheros, Aristotelis and Vogt, Michel R and Liconti, Davide and Yang, Chenyu and Cangan, Barnabas Gavin and Hinchet, Ronan J and others},
  journal={arXiv preprint arXiv:2504.04259},
  year={2025}
}

@inproceedings{mok2021measuring,
  title={Measuring the network performance of google cloud platform},
  author={Mok, Ricky KP and Zou, Hongyu and Yang, Rui and Koch, Tom and Katz-Bassett, Ethan and Claffy, Kimberly C},
  booktitle={Proceedings of the 21st ACM internet measurement conference},
  pages={54--61},
  year={2021}
}

@inproceedings{sfiligoi2020characterizing,
  title={Characterizing network paths in and out of the clouds},
  author={Sfiligoi, Igor and Graham, John and Wuerthwein, Frank},
  booktitle={EPJ Web of Conferences},
  volume={245},
  pages={07059},
  year={2020},
  organization={EDP Sciences}
}

@article{wang2025bitvla,
  title={BitVLA: 1-bit Vision-Language-Action Models for Robotics Manipulation},
  author={Wang, Hongyu and Xiong, Chuyan and Wang, Ruiping and Chen, Xilin},
  journal={arXiv preprint arXiv:2506.07530},
  year={2025}
}

\newpage
\appendix
\section{Detailed System and Model Parameters}
\label{sec:appendix:parameters}

In this section, we show the detailed hardware performance configuration used in our evaluation and the model parameters of $\pi_0$ in \Cref{{tab:pi0_model_specs}}.

\begin{table*}[h]
    \centering
    \setlength\dashlinedash{0.2pt}
    \setlength\dashlinegap{1.5pt}
    \begin{footnotesize}
    \caption{Hardware specifications for the GPUs used in our evaluation.}
    \label{tab:hardware_specs}
    \scalebox{1.0}{
        \begin{tabular}{L{6em} R{6em} R{6em} R{6em} R{4em} R{5em}}
\toprule
\textbf{Hardware}
& \multicolumn{1}{c}{\textbf{FP32}}
& \multicolumn{1}{c}{\textbf{BF16/FP16}}
& \multicolumn{1}{c}{\textbf{INT8}}
& \multicolumn{1}{c}{\textbf{Memory}}
& \multicolumn{1}{c}{\textbf{Memory BW}} \\
\midrule
Jetson Thor
& 100 TFLOP/s
& 400 TFLOP/s
& 800 TOP/s
& 128 GB
& 270 GB/s \\
\hdashline
RTX 4090
& 83 TFLOP/s
& 165 TFLOP/s
& 330 TOP/s
& 24 GB
& 1008 GB/s \\
\hdashline
A100
& 20 TFLOP/s
& 312 TFLOP/s
& 624 TOP/s
& 80 GB
& 2039 GB/s \\
\hdashline
H100
& 67 TFLOP/s
& 989 TFLOP/s
& 1979 TOP/s
& 80 GB
& 3350 GB/s \\
\hdashline
B100
& 60 TFLOP/s
& 1750 TFLOP/s
& 3500 TOP/s
& 192 GB
& 8000 GB/s \\
\bottomrule
        \end{tabular}
    }
    \end{footnotesize}
\end{table*}
\begin{table}[h]
    \centering
    \setlength\dashlinedash{0.2pt}
    \setlength\dashlinegap{1.5pt}
    \begin{footnotesize}
    \caption{Parameter specifications for $\pi_0$ model components (without vocabulary table).}
    \label{tab:pi0_model_specs}
    \scalebox{1.0}{
        \begin{tabular}{L{7em} R{3em} R{5em} R{5em} R{4em} R{4.5em} R{3.5em}}
\toprule
\textbf{Component}
& \multicolumn{1}{c}{\textbf{Layers}}
& \multicolumn{1}{c}{\textbf{Hidden Dim}}
& \multicolumn{1}{c}{\textbf{Interm. Dim}}
& \multicolumn{1}{c}{\textbf{Q Heads}}
& \multicolumn{1}{c}{\textbf{KV Heads}}
& \multicolumn{1}{c}{\textbf{Params}} \\
\midrule
Vision Encoder
& 27
& 1,152
& 4,304
& 16
& 16
& 411.19M \\
\hdashline
VLM Backbone
& 18
& 2,048
& 16,384
& 8
& 1
& 1.98B \\
\hdashline
Action Expert
& 18
& 1,024
& 4,096
& 8
& 1
& 292.63M \\
\bottomrule
        \end{tabular}
    }
    \end{footnotesize}
\end{table}

\end{document}